\documentclass[letterpaper,10pt,journal,twoside,romanappendices]{IEEEtran}  
\IEEEoverridecommandlockouts 

\usepackage{xparse}
\usepackage{color}
\usepackage{booktabs}
\usepackage{multirow}
\usepackage{multicol}
\usepackage{graphicx} 
\usepackage{comment}
\usepackage{hyperref}
\usepackage{array}
\usepackage{caption}
\usepackage{subcaption}
\usepackage{tikz}
\usetikzlibrary{positioning, arrows.meta}
\graphicspath{{figs/}}
\usepackage{float}
\usepackage{adjustbox}
\usepackage{overpic}
\usepackage{enumitem}
\usepackage{etoolbox}
\usepackage{colortbl} 
\usepackage{xcolor}
\usetikzlibrary{calc}

\usepackage{amsmath}
\usepackage{amssymb}
\usepackage{amsfonts}

\usepackage[ruled,vlined]{algorithm2e}
\usepackage{algpseudocode}
\SetArgSty{textnormal}      
\SetCommentSty{textnormal}

\usepackage{cite}	
\usepackage{balance}
\usepackage[capitalise]{cleveref}
\crefformat{equation}{(#2#1#3)}

\Crefname{figure}{Fig.}{Figs.}

\newcommand{\red}{\cellcolor{red!30}}
\newcommand{\orange}{\cellcolor{orange!30}}
\newcommand{\yellow}{\cellcolor{yellow!30}}
\newcommand{\xtimes}{\mathbin{\!\! \times \!\!}}

\title{A Photorealistic Dataset and Vision-Based Algorithm for Anomaly Detection During\\ Proximity Operations in Lunar Orbit}

\author{Selina Leveugle$^1$, Chang Won Lee$^2$, Svetlana Stolpner$^3$, Chris Langley$^3$,\\ Paul Grouchy$^3$, Steven Waslander$^2$, and Jonathan Kelly$^1$
\thanks{Manuscript received: August 20, 2025; Revised November 15, 2025; Accepted December 21, 2025.}
\thanks{This paper was recommended for publication by 
Giuseppe Loianno upon evaluation of the Associate Editor and Reviewers' comments.}
\thanks{$^1$Space and Terrestrial Autonomous Robotic Systems (STARS) Laboratory at the University of Toronto Institute for Aerospace Studies (UTIAS), Toronto, Canada. Email: \texttt{<firstname>.\allowbreak <lastname>\allowbreak@robotics.utias.utoronto.ca}}
\thanks{$^2$Toronto Robotics and AI Laboratory (TRAIL) at the University of Toronto Institute for Aerospace Studies (UTIAS), Toronto, Canada. Email: \texttt{<firstname>.\allowbreak<lastname>\allowbreak@robotics.utias.utoronto.ca}}
\thanks{$^3$MDA Space Inc. Email: \texttt{<firstname>.\allowbreak <lastname>\allowbreak@mda.space}}
\thanks{Digital Object Identifier (DOI): see top of this page.}
}

\begin{document}

\markboth{IEEE Robotics and Automation Letters. Preprint Version. Accepted December 2025}
{Leveugle \MakeLowercase{\textit{et al.}}: A Photorealistic Dataset and Vision-Based Algorithm for Anomaly Detection in Lunar Orbit} 

\maketitle
\bstctlcite{IEEEtranBSTCTL}

\begin{abstract}
NASA's forthcoming Lunar Gateway space station, which will be uncrewed most of the time, will need to operate with an unprecedented level of autonomy. One key challenge is enabling the Canadarm3, the Gateway's external robotic system, to detect hazards in its environment using its onboard inspection cameras. This task is complicated by the extreme and variable lighting conditions in space. In this paper, we introduce the visual anomaly detection and localization task for the space domain and establish a benchmark based on a synthetic dataset called ALLO (Anomaly Localization in Lunar Orbit). We show that state-of-the-art visual anomaly detection methods often fail in the space domain, motivating the need for new approaches. To address this, we propose MRAD (Model Reference Anomaly Detection), a statistical algorithm that leverages the known pose of the Canadarm3 and a CAD model of the Gateway to generate reference images of the expected scene appearance. Anomalies are then identified as deviations from this model-generated reference. On the ALLO dataset, MRAD surpasses state-of-the-art anomaly detection algorithms, achieving an AP score of 62.9\% at the pixel level and an AUROC score of 75.0\% at the image level. Given the low tolerance for risk in space operations and the lack of domain-specific data, we emphasize the need for novel, robust, and accurate anomaly detection methods to handle the challenging visual conditions found in lunar orbit and beyond.
\end{abstract}

\begin{IEEEkeywords}
Space Robotics and Automation, Data Sets for Robotic Vision, Simulation and Animation, Anomaly Detection.
\end{IEEEkeywords}

\section{Introduction}
\label{sec:intro}
\IEEEPARstart{O}{VER} the past two decades, the international space community has begun to focus its efforts on extending human space exploration beyond low-Earth orbit. 
NASA's Artemis program aims to deploy the Lunar Gateway, the first space station in lunar orbit, that will test new technologies needed for extended deep-space missions \cite{crusan_deep_2018}.
Unlike the International Space Station (ISS), the Gateway will be required to operate autonomously and without an on-board crew for long periods.
Complete autonomy is advantageous for the Canadarm3, the external robotic system being developed by MDA Space for the Canadian Space Agency. It will play a multifaceted role on the Gateway, including station maintenance, inspection, and capture and berthing of visiting vehicles \cite{agency_about_2020}.

To enhance robotic autonomy, vision-based anomaly detection methods that can  detect collision hazards, such as loose tools or debris, are of particular interest. 
Anomaly detection and localization refer to identifying and pinpointing image regions whose content deviates from the expected distribution of inputs. 
Anomaly detection in the space domain is particularly challenging because of complex lighting conditions that result from the black background coupled with harsh direct solar illumination.
Furthermore, the varied camera viewpoints used during operations near the space station can cause anomalies to fall into shadow regions and blend in with the background, making detection substantially more difficult.

\begin{figure}[tbp]
\vspace{1mm}
    \centering
	\begin{subfigure}{0.493\linewidth}
        \centering
        \begin{overpic}[width=\linewidth]{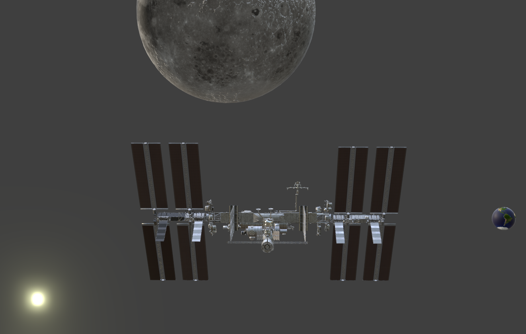}
            \put(85,55){\color{white}\bfseries \small (a)}
        \end{overpic}
    \end{subfigure}
    \hfill
	\begin{subfigure}{0.493\linewidth}
        \centering
        \begin{overpic}[width=\linewidth]{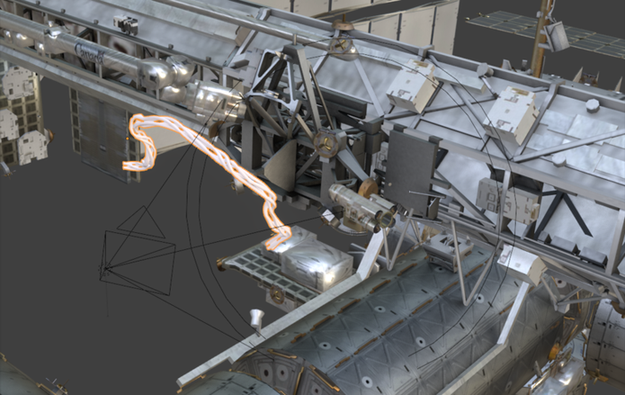}
            \put(85,55){\color{white}\bfseries \small (b)}
        \end{overpic}
    \end{subfigure}
    \\ \vspace{1mm} 
	\begin{subfigure}{0.493\linewidth}
        \centering
        \begin{overpic}[width=\linewidth]{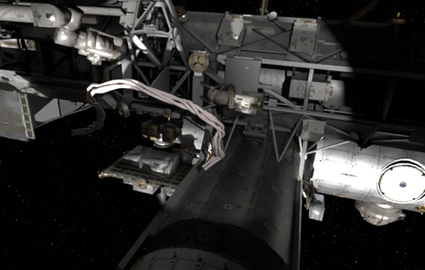}
            \put(85,55){\color{white}\bfseries \small (c)}
        \end{overpic}
    \end{subfigure}
    \hfill
	\begin{subfigure}{0.493\linewidth}
        \centering
        \begin{overpic}[width=\linewidth]{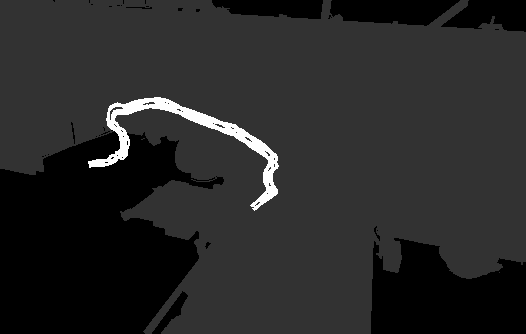}
            \put(85,55){\color{white}\bfseries \small (d)}
        \end{overpic}
    \end{subfigure}
    \caption{Visualization of the rendering process used to generate an anomalous image in the ALLO dataset. (a) Positions of the Sun, Moon, and space station are determined and scene lighting is configured accordingly; (b) An anomaly is inserted into the scene; 
    (c) The Blender Cycles engine renders the image; and (d) the corresponding three-class segmentation mask is generated.}
\label{fig:flow_chart}
\vspace{-4mm}
\end{figure}

Although reliable automated anomaly detection would be highly valuable, existing datasets and methods do not directly address the problem of anomaly detection in the space domain. 
Specifically, a significant challenge in space anomaly detection is the absence of comprehensive, labelled datasets necessary for robust evaluation. 
Moreover, our comparative analysis in this paper demonstrates that existing approaches significantly underperform, failing to generalize effectively to this domain.
To address these gaps, we introduce ALLO, a large-scale open-source anomaly detection dataset tailored for robotic proximity operations in space. Using ALLO, we establish a benchmark by evaluating existing anomaly segmentation algorithms. We further present MRAD, a novel statistical algorithm specifically designed for this application that leverages camera pose information and a known CAD model of the station to perform per-image anomaly segmentation, achieving state-of-the-art performance on the ALLO benchmark. 
In summary, our main contributions are as follows.
\begin{itemize}
\item We introduce ALLO, a synthetic dataset for vision-based anomaly detection in the space domain, and release its open-source generation pipeline to facilitate extensions and future research.
\item We establish an anomaly detection benchmark using ALLO, evaluating recent anomaly segmentation algorithms and highlighting their shortcomings for the space environment.
\item We introduce Model Reference Anomaly Detection (MRAD), a statistical algorithm designed for the space domain, and demonstrate its superior performance over existing anomaly detection algorithms on ALLO through comprehensive evaluation and ablation studies.
\end{itemize}

The remainder of the paper is structured as follows. In \Cref{sec:related work}, we review existing anomaly detection datasets and algorithms. In \Cref{sec:dataset}, we detail the ALLO dataset and its rendering pipeline. We then describe our proposed MRAD algorithm in \Cref{sec:methodology} and evaluate its performance against other state-of-the-art methods on the ALLO dataset in \Cref{sec:experiment}. The ALLO dataset and all open-source code are available at \url{https://github.com/utiasSTARS/ALLO}.

\section{Related Work}
\label{sec:related work}

Although anomaly detection is an emerging topic in the space domain, there is extensive work on visual datasets for space applications and, separately, on industrial anomaly detection. In this section, we review existing image datasets for these two applications. In addition, we provide an overview of modern anomaly detection methods that will serve as baselines for our benchmark.

\subsection{Datasets for Space Applications and Defect Inspection}
\label{subsec:dataset background}

\subsubsection{Space Applications}
Synthetic visual datasets are widely used in space applications due to the difficulty of acquiring real in situ imagery. Photorealistic rendering engines such as the Blender Cycles \cite{foundation_blenderorg_nodate} engine and Unreal Engine 5 \cite{noauthor_unreal_nodate} are commonly employed to produce simulated space imagery. Existing datasets typically focus on specific mission scenarios. For instance, Airbus’s SurRender \cite{brochard_scientific_2018} generates images for satellite servicing and debris removal, while the Space Imaging Simulator for Proximity Operations (SISPO) \cite{pajusalu_sispo_2022} leverages Blender to model asteroid fly-bys.

Physical simulators have also been developed. Legentil et al. \cite{legentil2025mixing}, for example, capture docking-port imagery under realistic dynamics and challenging lighting conditions to support satellite rendezvous and docking operations. Their dataset explicitly excludes the satellite body to ensure satellite-agnostic applicability for docking-port detection and state estimation.

In contrast, the ALLO dataset provides a comprehensive photorealistic simulation of an entire space station in lunar orbit, specifically designed for anomaly detection to mitigate collision risks. Its open-source data-generation pipeline further enables adaptation to diverse downstream tasks.

\subsubsection{Industrial Defect Inspection}

The majority of recent efforts in anomaly detection have focused on finding manufacturing defects during visual industrial inspection. The MVTec 2D anomaly detection dataset \cite{bergmann_mvtec_2019} is the most widely used benchmark for this application due to its pixel-level annotations and diverse range of objects \cite{zheng_benchmarking_2022}. However, similar to other anomaly detection datasets such as BTech \cite{btech_dataset} and Kolektor \cite{Tabernik2019JIM}, MVTec images are captured from fixed viewpoints, under consistent lighting and with simple backgrounds. In contrast, the ALLO dataset offers a more comprehensive and diverse set of scenes, including multiple views of the ISS with varied lighting conditions and complex backgrounds.

\vspace{-1mm}
\subsection{Anomaly Detection Methods}
\label{subsec:AD background}

Anomaly detection involves identifying abnormal samples, such as unexpected or irregular instances, that deviate from an expected distribution.
Existing methods fall broadly into two categories: statistical approaches, which rely on (possibly learned) statistical models, and deep learning-based approaches, which use deep neural networks to learn a representation of normal (non-anomalous) data.

\subsubsection{Statistical Detectors}
\label{subsec:trad AD}

Statistical detection methods compute anomaly scores using distance functions or statistical tests, followed by binary classification via empirical thresholds or probabilistic models \cite{ruffUnifyingReviewDeep2021a}. Classical techniques such as one-class support vector machines (OC-SVM) \cite{manevitzOneClassSVMsDocument2002} and support vector data descriptors (SVDD) \cite{hoffmannKernelPCANovelty2007a} model the density or decision boundary of normal data to identify outliers. 
The Reed–Xiaoli detector (RXD) \cite{reedAdaptiveMultiplebandCFAR1990}, originally developed for use with hyperspectral aerial imagery, assumes a multivariate Gaussian distribution for background pixels and detects anomalies as probabilistic outliers. However, this assumption fails for complex, non-homogeneous scenes such as those encountered in space.

Several RXD extensions relax its homogeneity requirement. The probabilistic anomaly detector (PAD) \cite{gaoProbabilisticAnomalyDetector2014} separates pixels into target and background sets and compares their RXD scores, while the random selection-based anomaly detector (RSAD) \cite{duRandomSelectionBasedAnomalyDetector2011} iteratively redefines the background set to improve robustness. Nevertheless, these methods remain sensitive to background variability and are not designed to exploit geometric priors available in space applications.

\subsubsection{Deep Learning-Based Detectors}
\label{subsubsec:learned AD}

Most modern deep learning-based anomaly detectors are unsupervised due to the scarcity of labelled anomalies, although some self-supervised approaches introduce synthetic anomalies during training \cite{zavrtanik_dsr_2022}. Unsupervised methods \cite{deng_anomaly_2022,lee_cfa_2022} learn a representation of normal data and detect deviations at inference time. Many operate in feature space, using pre-trained backbones (e.g., ResNet) and either fit statistical models to extracted features \cite{uninit_wei,lee_cfa_2022} or estimate their density with normalizing flows \cite{gudovskiy_cflow-ad_2021,yu_fastflow_2021}. 
In \cite{matteo_segmentation}, an encoder-decoder framework is combined with anomaly scores similar to RXD to identify pixels far from a Gaussian distribution.
Alternatively, generative approaches reconstruct only normal images, flagging anomalies by their reconstruction error \cite{zheng_benchmarking_2022}. Student–teacher architectures \cite{wang_student-teacher_2021, deng_anomaly_2022} follow this paradigm by distilling normal data representations from a teacher to a student, thereby failing to reproduce unseen anomalous patterns.

Although these methods achieve state-of-the-art results in structured domains such as industrial defect detection, they generalize poorly to the space domain due to high scene variability, complex lighting, and limited representative training data for learning.

\section{The ALLO Pipeline and Dataset}
\label{sec:dataset}

Our first contribution is a dataset for space-based anomaly detection and localization. The ALLO dataset comprises 51,409 synthetic images simulating the visual feed from arm-mounted cameras. The dataset is divided into a training set of normal (anomaly-free) images and a test set containing both normal and anomalous examples. Each image is generated with the Blender Cycles path-tracing engine, realistically simulating the lighting conditions in lunar orbit, including the blackness of deep space, direct sunlight, and inspection lighting.  The image-generation pipeline, described below, is fully open source and can be adapted for related computer vision tasks, such as semantic segmentation and depth estimation.

\subsection{Scene Modelling}
\label{subsec:model setup}

We use a high-fidelity Blender model of the ISS \cite{noauthor_international_nodate} as a stand-in for the Lunar Gateway, since a detailed Gateway model is currently unavailable.  Given the expected architectural and surface similarities of the Gateway to the ISS, this model provides a realistic approximation of the visual environment anticipated for the Canadarm3's operations.  The Gateway's anticipated orbit \cite{leeWhitePaperGateway2019} is approximated by an ellipse matching the expected perilune and apolune. 

The positions of the Earth and Sun are determined over a 365-day period relative to the Moon using 2030 ephemeris data from the Skyfield library \cite{2019ascl.soft07024R}. These bodies are positioned relative to the ISS model to replicate the background views expected by the arm's cameras. For computational efficiency, all models are scaled down in size but positioned to preserve their apparent angular sizes from the station's perspective, ensuring visual fidelity.

To simulate arm-mounted camera perspectives, 50 unique base poses are manually defined around the ISS. These poses are derived from possible Canadarm3 trajectories near the mounting fixtures located on the exterior of the station. To introduce realistic pose imprecision and enhance viewpoint diversity, each base pose is perturbed with Gaussian noise during rendering, using standard deviations of 1 m for position and 0.2 rad for orientation (roll, pitch, and yaw). This randomization strategy ensures comprehensive coverage of the key station regions and subtle yet critical variations in perspective.

\begin{figure}[t!]
\begin{subfigure}[b]{0.325\linewidth}
    \centering
    \includegraphics[width=\linewidth]{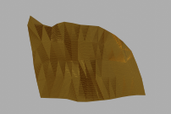}
    \caption{Thermal blanket}
    \label{fig:blanket}
\end{subfigure}
\hfill
\begin{subfigure}[b]{0.325\linewidth}
    \centering
    \includegraphics[width=\linewidth]{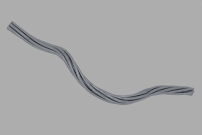}
    \caption{Cable}
    \label{fig:cable}
\end{subfigure}
\hfill
\begin{subfigure}[b]{0.325\linewidth}
    \centering
    \includegraphics[width=\linewidth]{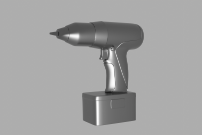}
    \caption{Drill}
    \label{fig:drill}
\end{subfigure}
\caption{Examples of models of anomalous objects used in the ALLO dataset.}
\label{fig:blender models}
\vspace{-4mm}
\end{figure}

\begin{figure*}[tbp]
\centering
\begin{subfigure}{0.195\textwidth}
    \includegraphics[width=\linewidth]{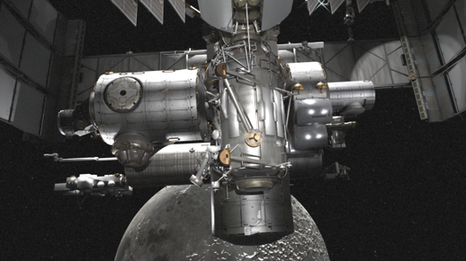}
\end{subfigure}
\begin{subfigure}{0.195\textwidth}
    \includegraphics[width=\linewidth]{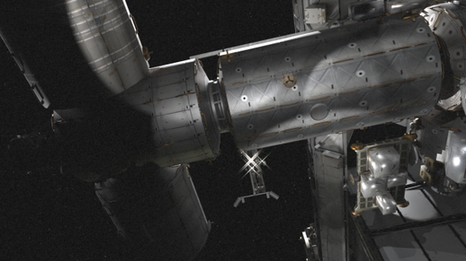}
\end{subfigure}
\begin{subfigure}{0.195\textwidth}
    \includegraphics[width=\linewidth]{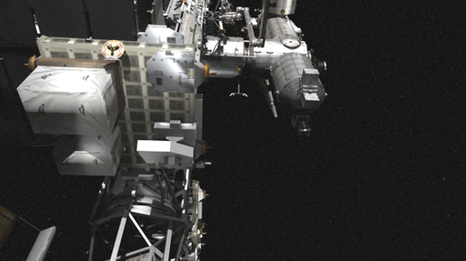}
\end{subfigure}
\begin{subfigure}{0.195\textwidth}
    \includegraphics[width=\linewidth]{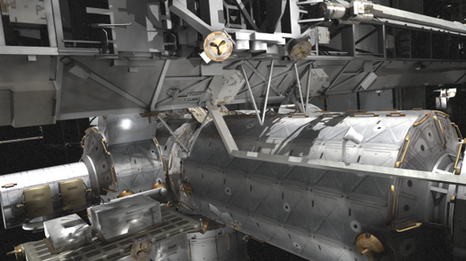}
\end{subfigure}
\begin{subfigure}{0.195\textwidth}
    \includegraphics[width=\linewidth]{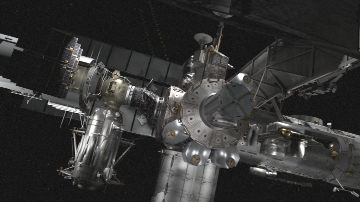}
\end{subfigure}
\\
\vspace{0.1cm}
\begin{subfigure}{0.195\textwidth}
    \includegraphics[width=\linewidth]{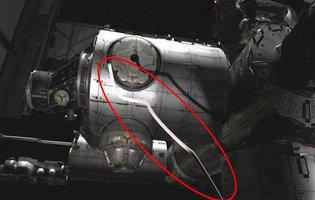}
\end{subfigure}
\begin{subfigure}{0.195\textwidth}
    \includegraphics[width=\linewidth]{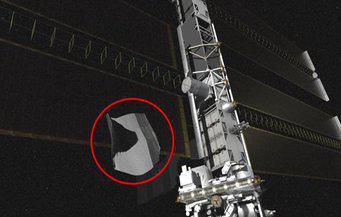}
\end{subfigure}
\begin{subfigure}{0.195\textwidth}
    \includegraphics[width=\linewidth]{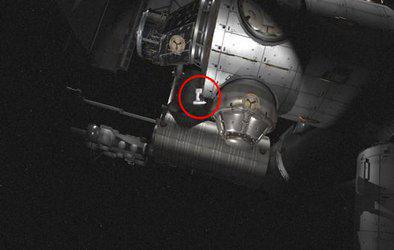}
\end{subfigure}
\begin{subfigure}{0.195\textwidth}
    \includegraphics[width=\linewidth]{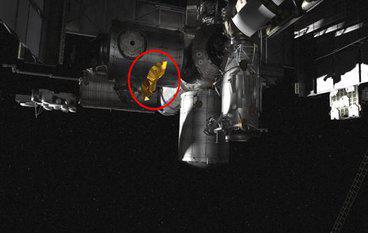}
\end{subfigure}
\begin{subfigure}{0.195\textwidth}
    \includegraphics[width=\linewidth]{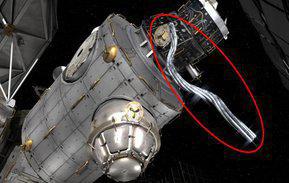}
\end{subfigure}
\caption{
    Sample images from the ALLO dataset. \textit{Top Row:} Normal (anomaly-free) images captured from five distinct camera views of the ISS. \textit{Bottom Row:} Images containing anomalies, identified by red ellipses, that include free-floating equipment such as cables, thermal blankets, and a hand drill.}
\label{fig:dataset_examples}
\vspace{-4mm}
\end{figure*}

\begin{figure}[tbp]
\centering
\begin{subfigure}{0.49\linewidth}
    \includegraphics[width=\linewidth]{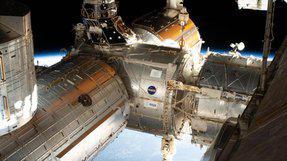}
\end{subfigure}
\begin{subfigure}{0.49\linewidth}
    \includegraphics[width=\linewidth]{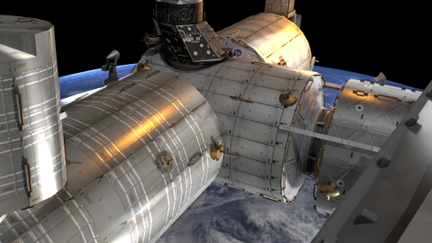}
\end{subfigure}
\\[3pt]
\begin{subfigure}{0.49\linewidth}
    \includegraphics[width=\linewidth]{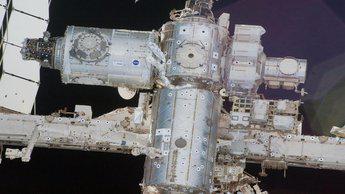}
\end{subfigure}
\begin{subfigure}{0.49\linewidth}
\includegraphics[width=\linewidth]{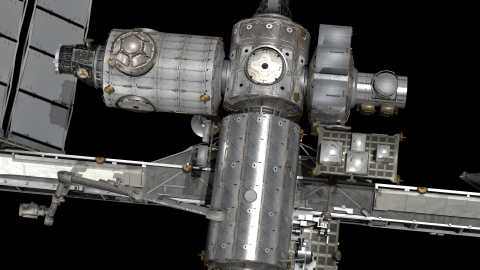}
\end{subfigure}
\caption{Real images of the ISS (left) and their synthetic recreations (right).}
\label{fig:iss_validation}
\vspace{-4mm}
\end{figure}

\subsection{Image Rendering}
\label{subsec:image rendering}

Different camera positions are used for the training and test sets to avoid overlap between views. The test set contains 16 distinct anomalous objects representing potential collision hazards (see \cref{fig:blender models}), including dislodged station equipment such as thermal blankets or cables, items lost during extravehicular activities such as tools or astronaut gloves, and objects resembling experimental equipment from prior missions such as Cubesats and small satellites. 

For each anomalous scene, only one such object is inserted into the camera’s view frustum to simplify evaluation. The object is randomly positioned within the station’s vicinity and rendered at five depths: one initial random depth and additional positions displaced by $\pm$1 m and $\pm$2 m from this initial position. If the object is occluded or projected to occupy less than 0.1\% of the image, it is repositioned to ensure visibility. This procedure yields a diverse set of anomaly appearances while maintaining realistic spatial consistency.

Each rendered image has a resolution of 1,920 $\xtimes$ 1,080 pixels and is paired with a three-class segmentation mask delineating the background, non-anomalous foreground (e.g., station surfaces, celestial bodies), and anomalous regions. The rendering pipeline, illustrated in \Cref{fig:flow_chart}, is repeated with two random seeds to expand the dataset. The dataset composition is reported in \Cref{tab:dataset images}, with sample images shown in \Cref{fig:dataset_examples}.

In summary, scenes were generated for each day in the ephemeris data by first positioning the Moon, Earth, and Sun according to their ephemerides. Camera poses were then selected, with indices 0--39 assigned to normal scenes and 40--49 reserved for anomalous ones. For each camera pose and anomaly configuration, the following steps were executed:
\begin{enumerate}[label=\arabic*)]
\item Place the camera with a Gaussian pose perturbation.
\item Add a nearby spotlight.
\item If the scene is anomalous:
\begin{enumerate}[label=\arabic{enumi}.\arabic*),nosep,leftmargin=2em]
\item Insert the anomaly into the view frustum.
\item Reposition the anomaly for visibility if necessary.
\end{enumerate}
\item Add noise and glare.
\item Render the image and corresponding segmentation mask using the Blender Cycles engine.
\end{enumerate}

\vspace{-1mm}
\subsection{Image Validation}
\label{subsec:allo validation}

To validate the realism of our synthetic ALLO images, we compared them with real photographs of the ISS from past ISS missions.
We matched Blender camera poses to these reference images, ensuring precise geometric alignment, producing synthetic renderings that closely replicate real lighting conditions and surface textures (see \Cref{fig:iss_validation}).

Quantitatively, we used the learned perceptual image patch similarity metric (LPIPS) \cite{zhangUnreasonableEffectivenessDeep2018} across ten real–synthetic pairs, obtaining an average score of 0.58---comparable to state-of-the-art image synthesis models such as GILL ($\sim$0.70 on VIST) \cite{kohGeneratingImagesMultimodal2023} and Stable Diffusion XL (0.88 on COCO2017) \cite{podellSDXLImprovingLatent2023}.
We also calculated the Fr\'{e}chet inception distance (FID) \cite{heusel2017gans} between 319 ALLO training images and historical ISS photographs, achieving a score of 3.61. This FID score surpasses those of other recent models, such as DiGIT \cite{zhu2024stabilize} (4.59 on ImageNet) and LC-GAN \cite{lee2024linearly} (5.77 on AFHQ-v2). Together, these two metrics confirm the high quality and photorealism of our synthetic images.

Together, these qualitative and quantitative results confirm that the synthetic images effectively reproduce the visual characteristics of the target space environment.

\begin{table}[b]
\vspace{-2mm}
\caption{Composition of the ALLO dataset (camera poses and images) across the training and test sets. The training set is composed exclusively of normal images, while the test set contains both normal and anomalous images rendered from disjoint sets of camera poses.}
\label{tab:dataset images}
\centering
\renewcommand{\arraystretch}{1.1}
\begin{tabular}{ l c c c}
\toprule
  \textbf{Dataset Set} & \textbf{Training} & \textbf{Test} & \textbf{Total}\\ \midrule
Camera Poses & 1--40 & 41--50 & 50 \\ 
Normal Images & 29,085 & 7,300 & 36,385 \\
Anomalous Images & 0 & 15,024 & 15,024\\
\bottomrule
\end{tabular}
\end{table}

\section{Model Reference Anomaly Detection}
\label{sec:methodology}

MRAD is an anomaly detection algorithm designed for the space domain. It detects anomalies by comparing a query image with a reference image rendered from a known camera pose and a CAD model of the station. The algorithm operates in three stages: (1) pixel-level anomaly scores are computed using an extended version of the Reed–Xiaoli detector (RXD) applied to query–reference pairs; (2) region growing clusters high-scoring pixels into candidate anomalous regions, producing a binary mask; and (3) an image is classified as anomalous if the total detected area exceeds a threshold. In this work, both reference and query images are synthetically generated, and the real-to-synthetic domain gap is simulated by introducing random variations in lighting and small perturbations to the camera pose when rendering the reference images.

\vspace{-1mm}
\subsection{Reed-Xiaoli Detector (RXD) Anomaly Score}
\label{subsec: rxd scores}

Each pixel in the query image is first assigned an anomaly score based on the Reed–Xiaoli detector (RXD) \cite{reedAdaptiveMultiplebandCFAR1990}, which measures a pixel’s statistical distance from the background.
RXD models nominal pixels as samples from a multivariate Gaussian distribution with mean vector $\boldsymbol{\mu} \in \mathbb{R}^3$ and covariance matrix $\boldsymbol{\Sigma} \in \mathbb{R}^{3\times3}$. Pixels with low likelihood under $\mathcal{N}(\boldsymbol{\mu}, \boldsymbol{\Sigma})$ are considered anomalous. Given an image $I$ with mean $\boldsymbol{\mu}_I$ and covariance $\boldsymbol{\Sigma}_I$, the RXD anomaly score for a pixel $\mathbf{x}$ is the squared Mahalanobis distance:
\begin{equation}\label{eq:rxd}
\mathrm{RXD}_{I}(\mathbf{x}) = (\mathbf{x}-\boldsymbol{\mu}_I)^{T}\boldsymbol{\Sigma}_I^{-1}(\mathbf{x}-\boldsymbol{\mu}_I).
\end{equation}

While effective for relatively homogeneous scenes, RXD struggles with non-uniform backgrounds, such as space imagery, where lighting and texture vary significantly.
To address this limitation, we adapt principles from the probabilistic anomaly detector (PAD) \cite{gaoProbabilisticAnomalyDetector2014}, which compares likelihoods under both nominal and anomalous distributions. In MRAD, the nominal (anomaly-free) distribution is derived from the reference image, while the query image provides statistics for potentially anomalous regions. Specifically, the anomaly-free distribution statistics $(\boldsymbol{\mu}_0,\boldsymbol{\Sigma}_0)$ are computed from the reference image, and the query distribution statistics $(\boldsymbol{\mu}_1,\boldsymbol{\Sigma}_1)$ are computed from the query image. By leveraging statistics from both the reference and query images, MRAD can better handle complex, varying illumination and background structure, where the pixel distribution is non-Gaussian. The MRAD anomaly score for a pixel $\mathbf{x}$ is then defined as:
\begin{equation}
\label{eq:mpad score}
\mathrm{MRAD}(\mathbf{x}) = \mathrm{RXD}_{\mathrm{ref}}(\mathbf{x}; \boldsymbol{\mu}_0, \boldsymbol{\Sigma}_0) - \mathrm{RXD}_{\mathrm{query}}(\mathbf{x}; \boldsymbol{\mu}_1, \boldsymbol{\Sigma}_1),
\end{equation}
where, given nominal image alignment, $\mathrm{RXD}_{\mathrm{ref}}(\mathbf{x})$ is expected to be greater than or equal to $\mathrm{RXD}_{\mathrm{query}}(\mathbf{x})$.

\begin{figure}[tbp]
\centering
\begin{tikzpicture}

\node[anchor=south west, inner sep=0] (img1) at (-0.1,0)
{\includegraphics[width=2.1cm]{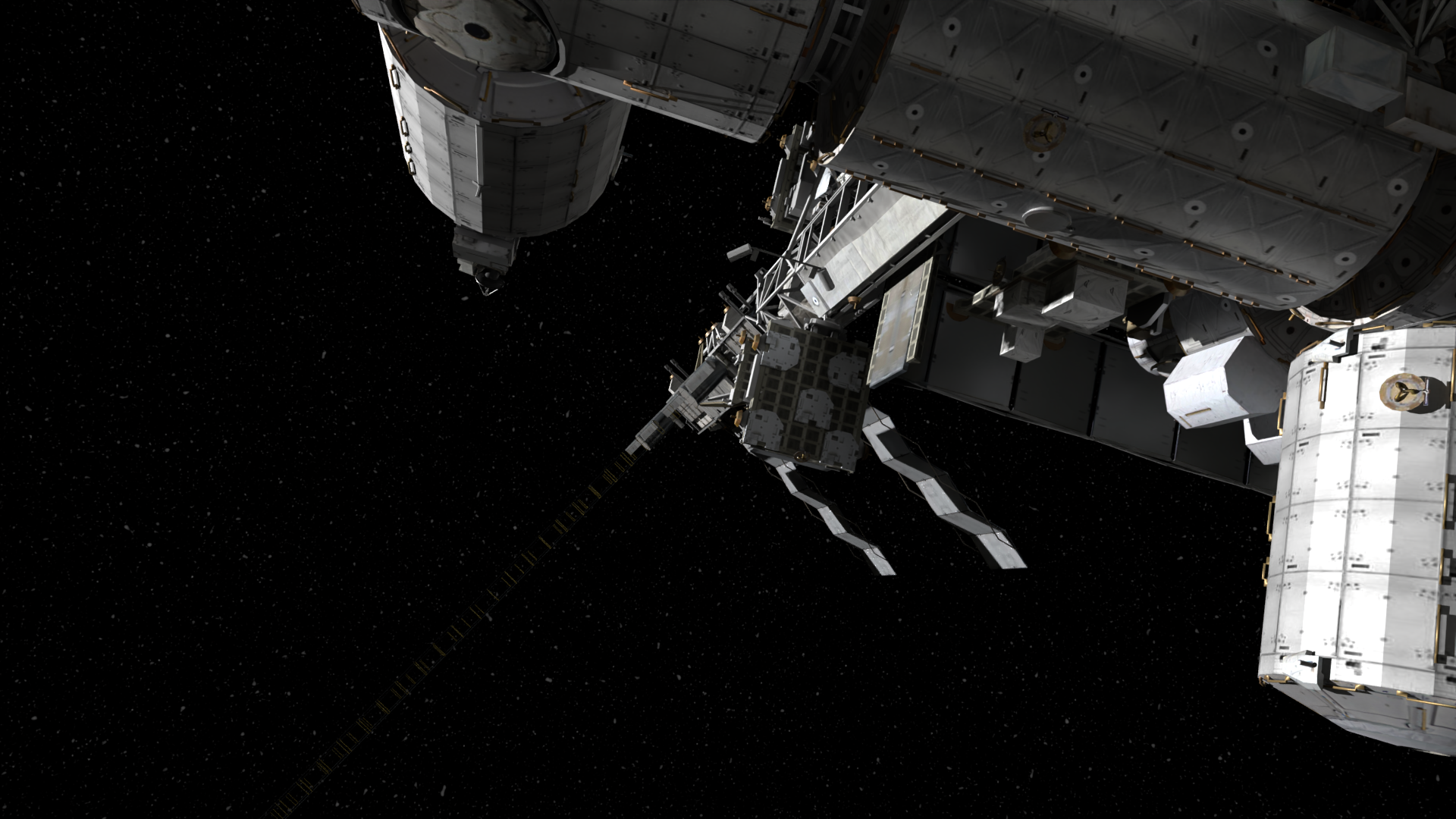}};

\begin{scope}

    \foreach \x in {0.25,0.5,0.75} {
        \draw[blue, line width=0.6pt]
            ($ (img1.south west)! \x ! (img1.south east) $) --
            ($ (img1.north west)! \x ! (img1.north east) $);
    }

    \foreach \y in {0.25,0.5,0.75} {
        \draw[blue, line width=0.6pt]
            ($ (img1.south west)! \y ! (img1.north west) $) --
            ($ (img1.south east)! \y ! (img1.north east) $);
    }

    \coordinate (cellBL) at
        ($ (img1.south west)!1.5!(img1.south east)!0.5!(img1.north west) $); 
    \coordinate (cellTR) at
        ($ (img1.south west)!2.0!(img1.south east)!0.75!(img1.north west) $); 

    \draw[red, very thick] (cellBL) rectangle (cellTR);

    \coordinate (tile23tr) at
        ($ (cellTR) + (0.5,0) $); 
    \coordinate (tile23br) at
        ($ (cellBL)!1!(cellTR|-cellBL) + (0.5,0) $); 

\end{scope}

\node[anchor=south west, inner sep=0, draw=black, thick] (img2) at (2.2,0)
{\includegraphics[width=1.75cm]{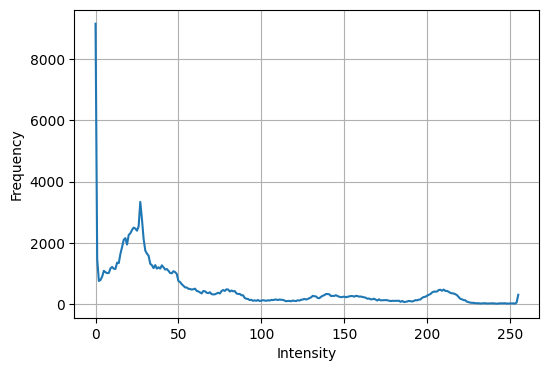}};

\draw[red, thick] (tile23tr) -- (img2.north west);
\draw[red, thick] (tile23br) -- (img2.south west);

\node[anchor=south west, inner sep=0] (img3) at (4.1,0)
{\includegraphics[width=2.1cm]{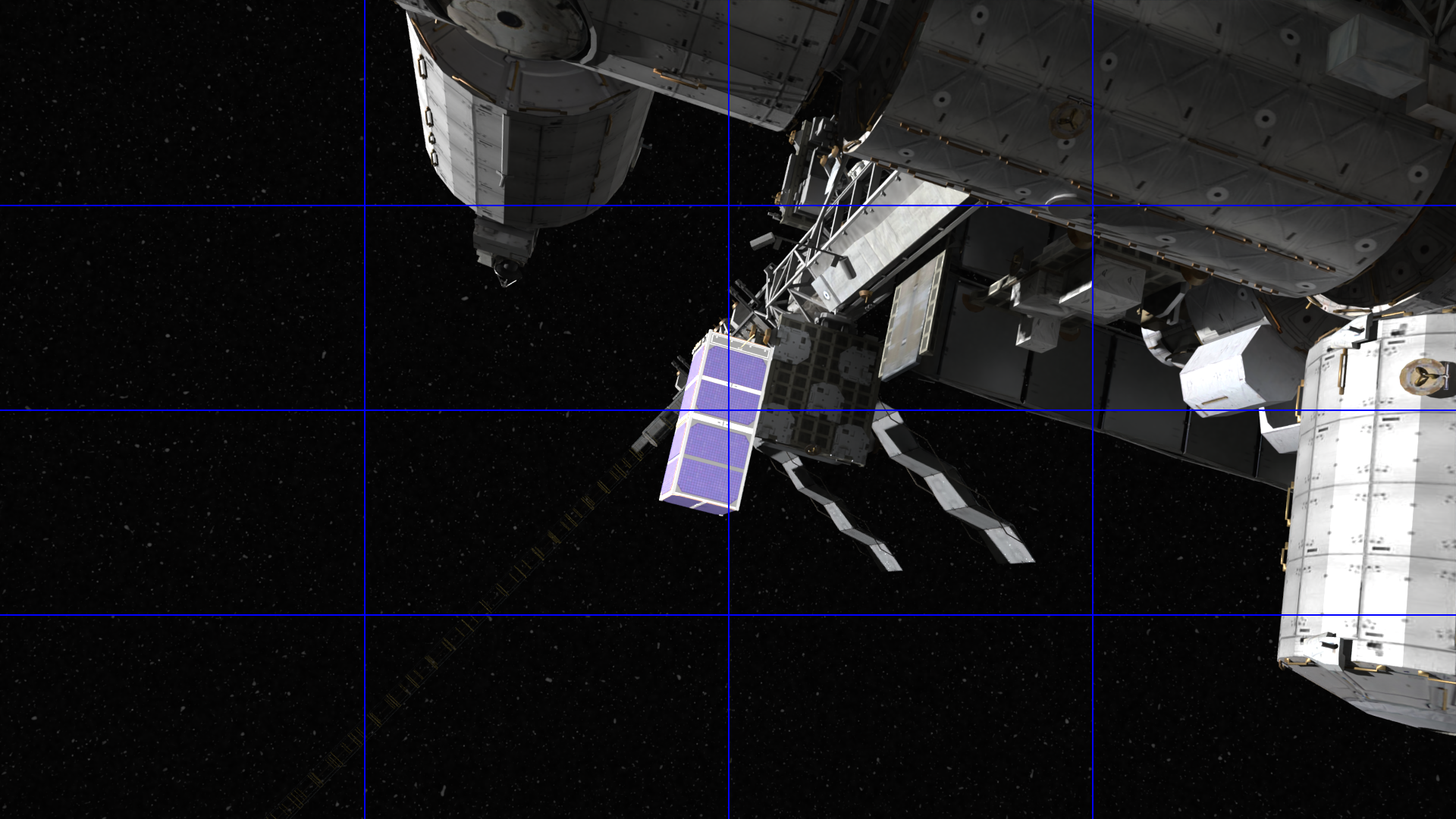}};

\node[anchor=south west, inner sep=0] (img4) at (6.3,0)
{\includegraphics[width=2.1cm]{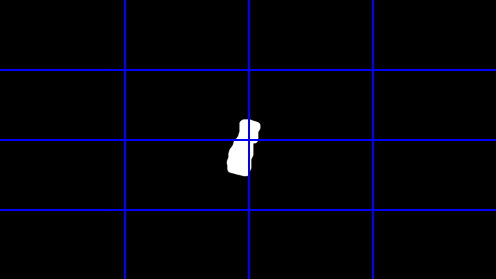}};

\end{tikzpicture}

\caption{Left to right: reference image from the ALLO test set with one  grid cell highlighted; pixel intensity histogram of the highlighted cell; anomalous query image; predicted anomaly map.}
\label{rebfig:pixel_analysis_combined}
\vspace{-4mm}
\end{figure}

To account for spatial variability in lighting and texture (e.g., the contrast between the black background of space and the station’s surface), both reference and query images are divided into 4 $\xtimes$ 4 grids. Mean and covariance statistics are computed per grid, yielding localized anomaly scores that better model region-specific variations. Scores are scaled in the range [0, 255] to produce a grayscale MRAD map, where pixels with values below a predefined threshold are classified as anomaly-free and set to zero, and remaining high-scoring pixels are retained for further analysis.
\Cref{rebfig:pixel_analysis_combined} illustrates a case where the pixel-intensity histogram for a reference grid cell is noticeably skewed and non-Gaussian, violating the Gaussian noise assumption. Nevertheless, in this instance MRAD remains robust, producing an accurate anomaly map that correctly localizes the free-floating cubesat.

\vspace{-1mm}
\subsection{Noise Removal}
\label{subsec: noise_removal}

A major source of false positives is random amplification in black background regions, where small color variations can produce spuriously high anomaly scores that are not effectively filtered by simple thresholding. To mitigate this issue, we analyze the spatial distribution of anomaly scores within each grid cell before generating the final anomaly score map.

First, a preliminary check identifies grid cells that are likely to contain a large anomaly: if more than 90\% of the scores in a cell exceed the minimum MRAD score, the entire cell is marked as anomalous.
Next, a kernel-based method removes noise caused by random score variations. A sliding-window kernel (with a size  $\sim$20\% of the grid dimensions) is used to compute the local standard deviation of anomaly scores; regions with a standard deviation below an empirical threshold (about 3–5\%) are considered to be indicative of uniform background noise and their scores are set to zero.
Finally, a Gaussian blur is applied to the noise-filtered score map to smooth local fluctuations and merge fragmented anomalous regions, improving region contiguity for subsequent detection.

\vspace{-1mm}
\begin{algorithm}[b!]
\caption{Model-Reference Anomaly Detection}
\label{alg:mpa}

\texttt{ScoreImage} $\leftarrow$ Zeros(SizeOf(\texttt{QueryImage}))\;
\texttt{AnomalyMap} $\leftarrow$ Zeros(SizeOf(\texttt{QueryImage}))\;

\ForEach{\texttt{GridCell} in \texttt{QueryImage}}{
  \ForEach{\texttt{Pixel} in \texttt{GridCell}}{
    \texttt{Score} $\leftarrow$ MRAD(\texttt{Pixel})$\!\!$ \tcp*[r]{See Eq.~(2)}
    \If{\texttt{Score} $>$ \texttt{MinMRADScore}}{
      \texttt{ScoreImage}[\texttt{Pixel}] $\leftarrow$ \texttt{Score}\;
    }
  }
  \texttt{StdDev} $\leftarrow$ StdDeviation(\texttt{GridCell}, \texttt{Kernel})\;
  \If{Range(\texttt{StdDev}) $<$ 3\%}{
    \texttt{Score} = 0 for all pixels in \texttt{GridCell}\;
  }
}

Apply Gaussian blur to \texttt{ScoreImage}\;
Apply mean shift clustering to \texttt{ScoreImage}\;
\texttt{Seeds} $\leftarrow$ FindClusterCentroids(\texttt{ScoreImage})\;

\ForEach{\texttt{StartPoint} in \texttt{Seeds}}{
  \texttt{AnomalyMap}[\texttt{StartPoint}] $\leftarrow $ 1$\!\!$ \tcp*[r]{Anomalous}
  \ForEach{\texttt{Pixel} in Footprint(\texttt{StartPoint})}{
    \If{InBounds(\texttt{Pixel}, \texttt{Tolerance})}{
      \texttt{AnomalyMap}[\texttt{Pixel}] $\leftarrow$ 1$\!\!$ \tcp*[r]{Anomalous}
    }
  }
}

\texttt{Count} $\leftarrow$ Sum(\texttt{AnomalyMap})\;
\If{\texttt{Count} $>$ \texttt{AnomalyThreshold}}{
  \texttt{ImageClass} $\leftarrow$ \texttt{Anomalous}\;
}
\Else{
  \texttt{ImageClass} $\leftarrow$ \texttt{Normal}\;
}
\end{algorithm}

\vspace{-1mm}
\subsection{Clustering via Region Growing}
\label{subsec:region growing}

Region growing \cite{adamsSeededRegionGrowing1994} is employed to cluster anomalous pixels into distinct objects based on pixel location and anomaly score. The implementation in MRAD relies on three parameters: the seed (initial pixel for expansion), the footprint (square search neighbourhood), and the tolerance (allowable score deviation for aggregation). Both the footprint and the tolerance are set empirically. 

Seed pixels are found by applying mean-shift clustering \cite{chengMeanShiftMode1995} to the blurred score map. This non-parametric, mode-seeking algorithm detects clusters by shifting each pixel toward the region of highest density, iterating until convergence at a local maximum. The resulting cluster centroids serve as seed points for region growing.

\begin{figure*}[tbp]
	\centering
	\begin{subfigure}{0.195\textwidth}
		\includegraphics[width=\linewidth]{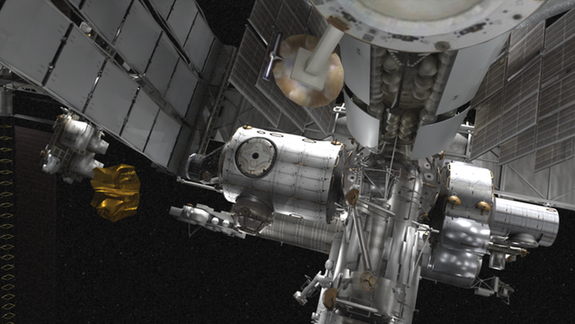}
		\subcaption{Query Image}
		\label{fig:mpa demo query}
	\end{subfigure}
	\begin{subfigure}{0.195\textwidth}
		\includegraphics[width=\linewidth]{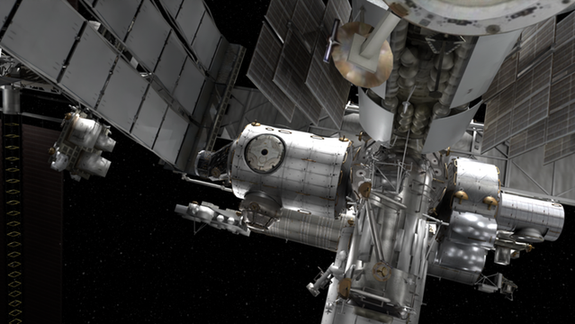}
		\subcaption{Reference Image}
		\label{fig:mpa demo ref}
	\end{subfigure}
	\begin{subfigure}{0.195\textwidth}
		\includegraphics[width=\linewidth]{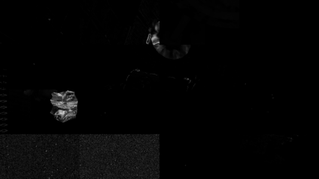}
		\subcaption{Anomaly Scores}
		\label{fig:mpa demo scores}
	\end{subfigure}
	\begin{subfigure}{0.195\textwidth}
		\includegraphics[width=\linewidth]{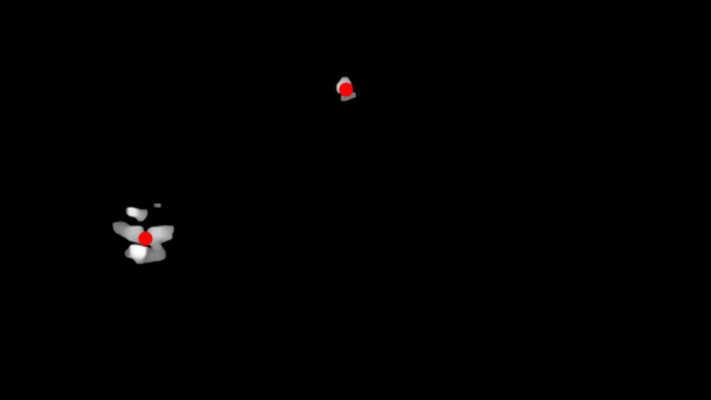}
		\subcaption{Region Growing Seeds}
		\label{fig:mpa demo blurred}
	\end{subfigure}
	\begin{subfigure}{0.195\textwidth}
		\includegraphics[width=\linewidth]{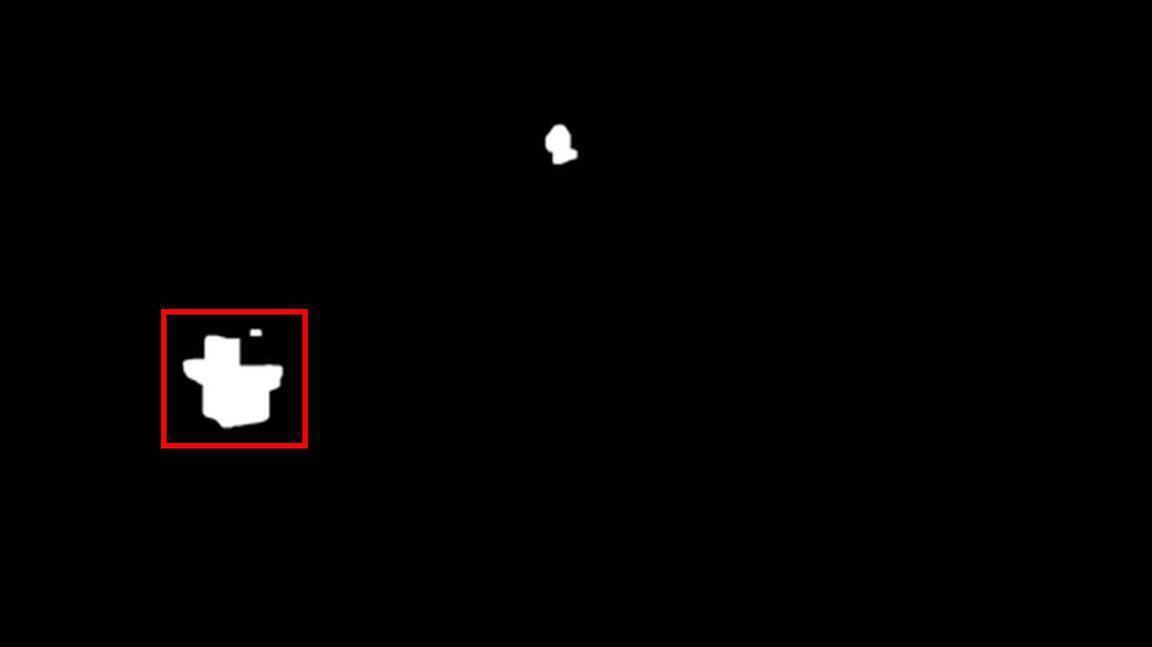}
		\subcaption{Anomaly Map}
		\label{fig:mpa demo map}
	\end{subfigure}
	\caption{Visualization of the steps in the MRAD algorithm. The query image (a) and reference image (b) are the algorithm inputs. The anomaly score image (c), computed using \cref{eq:mpad score}, highlights the anomalous thermal blanket. Noise is removed and a Gaussian blur is applied, followed by mean-shift clustering to identify region growing seed points (d). Region growing aggregates anomalous pixels to generate the final anomaly map (e), detecting the blanket while rejecting false positives. Notably, camera pose variations may introduce additional image differences that could be mistaken for anomalies if detection relied solely on pixel differences alone.}
    \label{fig:mpa demo}
\vspace{-5mm}
\end{figure*}

Region growing then expands from each seed, aggregating spatially adjacent pixels with similar anomaly scores until no further candidates meet the criteria. This yields a binary anomaly map, from which the total number of anomalous pixels is computed. An image is classified as anomalous if this count exceeds a predefined threshold. The complete MRAD pipeline is outlined in \cref{alg:mpa} and \cref{fig:mpa demo}.

\section{Experiments}
\label{sec:experiment}

Using the ALLO benchmark, we evaluate our proposed algorithm, MRAD, against recent unsupervised anomaly detection algorithms from the industrial and medical domains. We then analyze the limitations of these competing methods and investigate the factors that contribute to the superior performance of MRAD.

\subsection{Experimental Setup}
\label{subsec:setup}

We modified the Intel Anomalib library \cite{akcay_anomalib_2022} to load, train, and evaluate learning-based algorithms on the ALLO dataset.
We selected nine representative learning-based methods from the library, chosen to cover both reconstruction- and representation-based approaches and for their established performance on the MVTec industrial benchmark.

Normalizing flow-based methods, FastFlow \cite{yu_fastflow_2021} and UFlow \cite{tailanianUFlowUShapedNormalizing2024a}, learn the probability distribution of normal image features and detect anomalies as samples with low likelihood.
Student–teacher methods, including STFPM \cite{wang_student-teacher_2021}, Reverse Distillation \cite{deng_anomaly_2022} and UniNet \cite{uninit_wei} train a smaller student network to mimic the feature representations of a pre-trained teacher, flagging deviations as anomalies. UniNet introduces domain-aware feature selection, contrastive and margin losses, and a bottleneck with a weighted decision mechanism to support one model across multiple object categories without per-category training.
Self-supervised and reconstruction-based methods, such as DSR \cite{zavrtanik_dsr_2022}, SuperSimpleNet \cite{rolih2025supersimplenet}, and Dinomaly \cite{Guo_2025_CVPR}, reconstruct normal images. Some approaches leverage synthetic anomalies for supervision, whereas Dinomaly uses a minimalist pure‑Transformer reconstructor.
Finally, CFA \cite{lee_cfa_2022} is a feature-adaptation method that maps normal features to a shared hypersphere, detecting anomalies as features that fail to map correctly.
Representation-based methods generally use fixed encoders pre-trained on ImageNet \cite{krizhevskyImageNetClassificationDeep2012} and train remaining layers from scratch on the target dataset.

The proposed MRAD algorithm and the statistical detectors RXD and PAD were also evaluated on the ALLO test set with additional rendered reference images. RXD operates solely on the query image, while PAD and MRAD use both query and reference images; their anomaly scores were thresholded to produce output binary anomaly maps.
Ablation studies evaluated the impact of grid size, region-growing tolerance, kernel size, and the standard deviation used for noise removal on MRAD's detection performance. Parameters were varied around their empirically optimized values to assess robustness.

Performance was measured using image-level AUROC (I.AUROC), pixel-level AUROC (P.AUROC), pixel-level average precision (P.AP), and the pixel-level false-positive rate at 95\% true-positive rate (P.FPR95). On the imbalanced ALLO dataset, P.AUROC can be dominated by the negative class, obscuring localization failures. P.AP, which measures the precision-recall trade-off specifically for anomalous pixels, provides a more informative localization metric in this setting \cite{gudovskiy_cflow-ad_2021}. Consequently, P.AP was used as the primary metric for fine-grained anomaly segmentation, followed by P.FPR95.

\subsection{ALLO Benchmark Results}
\label{subsec:results}

The performance of existing learning-based algorithms and MRAD on the ALLO test set is summarized in \cref{tab:main results}.
Among the learning-based methods, Dinomaly achieves the highest P.AP of 49.4\%, followed by FastFlow (29.1\%) and UFlow (20.7\%), and also attains the best I.AUROC and P.AUROC on ALLO with scores of 78.6\% and 91.4\%, respectively. However, its performance dropped substantially relative to MVTec, with I.AUROC decreasing by 21.0 points (99.6\% to 78.6\%) and P.AUROC by 7.0 points (98.4\% to 91.4\%).

\begin{figure*}[tbp]
\centering
\begin{subfigure}{0.195\textwidth}
    \caption*{Query Image}
    \includegraphics[width=\linewidth]{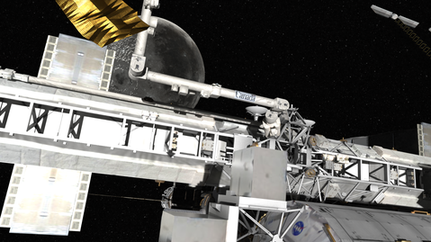}
\end{subfigure}
\begin{subfigure}{0.195\textwidth}
    \caption*{Ground Truth}
    \includegraphics[width=\linewidth]{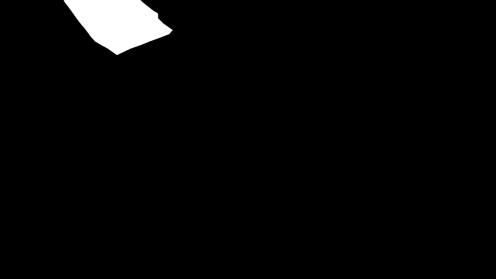}
\end{subfigure}
\begin{subfigure}{0.195\textwidth}
    \caption*{FastFlow Mask}
    \includegraphics[width=\linewidth]{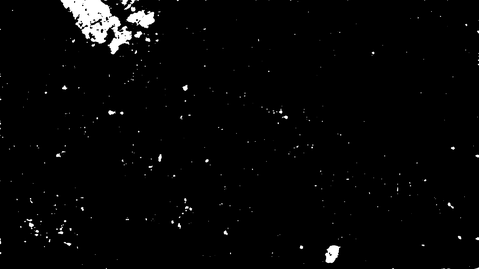}
\end{subfigure}
\begin{subfigure}{0.195\textwidth}
    \caption*{STFPM Mask}
    \includegraphics[width=\linewidth]{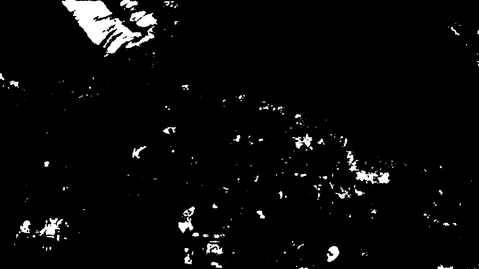}
\end{subfigure}
\begin{subfigure}{0.195\textwidth}
    \caption*{MRAD Mask}
    \includegraphics[width=\linewidth]{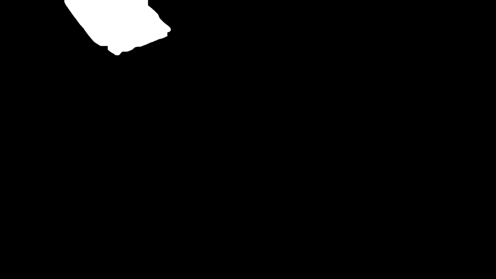}
\end{subfigure}
\\
\vspace{0.1cm}
\begin{subfigure}{0.195\textwidth}
    \includegraphics[width=\linewidth]{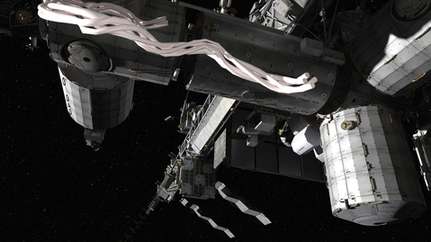}
\end{subfigure}
\begin{subfigure}{0.195\textwidth}
    \includegraphics[width=\linewidth]{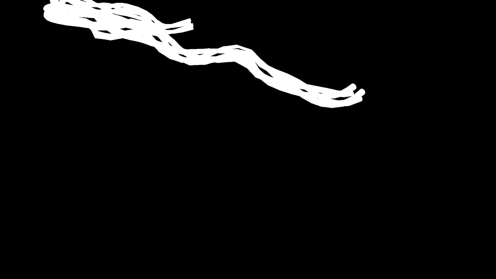}
\end{subfigure}
\begin{subfigure}{0.195\textwidth}
    \includegraphics[width=\linewidth]{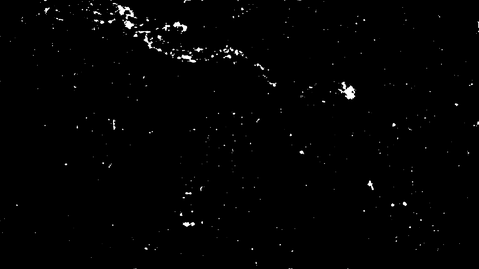}
\end{subfigure}
\begin{subfigure}{0.195\textwidth}
    \includegraphics[width=\linewidth]{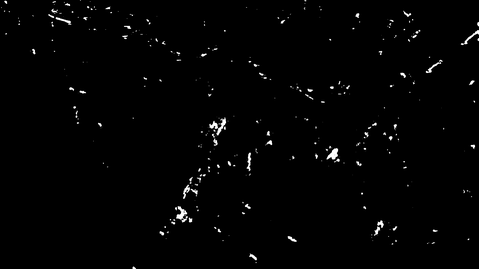}
\end{subfigure}
\begin{subfigure}{0.195\textwidth}
    \includegraphics[width=\linewidth]{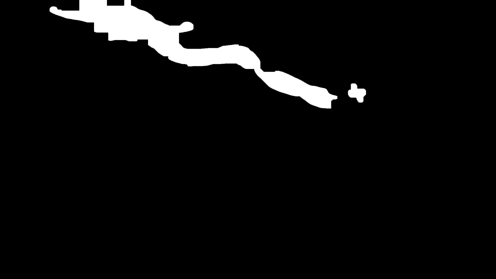}
\end{subfigure}
\caption{Examples of predicted anomaly masks for two images from the ALLO test set by two learning-based algorithms, FastFlow and STFPM, and by MRAD. Ground truth masks are provided for comparison. \textit{Top Row:} The anomalous object, a thermal blanket, is located in the upper left of the query image. \textit{Bottom Row:} The anomalous object, a power cable, is located in the upper centre of the query image.}
\label{fig:default inference}
\vspace{-4mm}
\end{figure*}

In contrast, MRAD, our learning-free method, outperforms all learning-based and learning-free baselines on ALLO (see \cref{tab:mrad_allo_results}), achieving the highest P.AP of 62.9\%, best FPR95 of 37.5\%, and the second-highest I.AUROC of 75.0\%. These results demonstrate that the grid-based dual statistical framework of MRAD enables more accurate detection of fine-grained anomalies. 
Qualitative examples of FastFlow, STFPM and MRAD are shown in \cref{fig:default inference}.

\setlength{\tabcolsep}{5pt}
\begin{table*}[tbp]
\setlength{\fboxsep}{2pt}
\newcommand{\smallstrut}{\rule[-0.4ex]{0pt}{0.9em}} 
 \caption{Performance of state-of-the-art anomaly detection algorithms on the ALLO test set. For each metric, the three top-performing methods are highlighted in \colorbox{red!30}{\smallstrut red}, \colorbox{orange!30}{\smallstrut orange} and \colorbox{yellow!30}{\smallstrut yellow}, respectively.}
    \label{tab:main results}
    \centering
    \begin{tabular}{c*{9}{c}|c}
        \toprule
        Metric 
        & \shortstack{DSR \\ \cite{zavrtanik_dsr_2022}} 
        & \shortstack{Rev. Dist. \\ \cite{deng_anomaly_2022}} 
        & \shortstack{SuperSimpleNet \\ \cite{rolih2025supersimplenet}}
        & \shortstack{CFA \\ \cite{lee_cfa_2022}} 
        & \shortstack{Dinomaly \\ \cite{Guo_2025_CVPR}} 
        & \shortstack{UniNet \\ \cite{uninit_wei}} 
        & \shortstack{STFPM \\ \cite{wang_student-teacher_2021}} 
        & \shortstack{UFlow \\ \cite{tailanianUFlowUShapedNormalizing2024a}} 
        & \shortstack{FastFlow \\ \cite{yu_fastflow_2021}} 
        & \shortstack{\textbf{MRAD} \\ (ours)} \\
        \midrule
        I.AUROC $\uparrow$  & 55.6 $\pm$ 2.0 & 57.4 $\pm$ 4.4 & 49.6 $\pm$ 0.5 & 51.3 $\pm$ 0.5 & \red 78.6 $\pm$ 0.5 & 56.8 $\pm$ 2.9 & 61.8 $\pm$ 2.4 & 60.7 $\pm$ 3.2 & \yellow 65.9 $\pm$ 0.6 & \orange 75.0\\
        P.AUROC $\uparrow$  & 69.5 $\pm$ 6.5 & 75.7 $\pm$ 0.7 & 79.5 $\pm$ 9.5 & 84.8 $\pm$ 7.7 & \red 91.4 $\pm$ 0.1 & \yellow 85.9 $\pm$ 2.2 & 87.8 $\pm$ 1.7 & 85.8 $\pm$ 1.6 & \orange 90.4 $\pm$ 0.5 & 71.8\\
        P.AP $\uparrow$ & 6.9 $\pm$ 0.7 & 7.0 $\pm$ 2.5 & 9.3 $\pm$ 1.0 & 11.0 $\pm$ 2.8 & \orange 49.4 $\pm$ 1.0 & 12.3 $\pm$ 2.4 & 12.5$\pm$ 1.3 & 20.7 $\pm$ 1.6 & \yellow 29.1 $\pm$ 1.2 & \red 62.9\\
        P.FPR95 $\downarrow$ & 95.4$\pm$2.5 & 95.0$\pm$1.7 & 64.4$\pm$18.3 & 58.7$\pm$29.1 & 47.7 $\pm$ 0.8 & 51.1 $\pm$ 13.6 & \yellow 43.0$\pm$7.0 & 66.7$\pm$10.6 & \orange 41.7$\pm$1.4 & \red 37.5 \\
        \bottomrule
    \end{tabular}
\vspace{-3mm}
\end{table*}

\begin{table}[tbp]
\caption{Performance of MRAD with ablations and of non-learning algorithms on the ALLO test set. The optimized settings are a 4 $\xtimes$ 4 grid, a tolerance of 50 MRAD score, kernel size of 20\%, and a SD of 5.}
\label{tab:mrad_allo_results}
\centering
\footnotesize
\resizebox{\columnwidth}{!}{
\begin{tabular}{l l c c c}
    \toprule
    \textbf{Method} & \textbf{Ablation} & \textbf{I. AUROC $\uparrow$} & \textbf{P. AUROC $\uparrow$} & \textbf{P. AP $\uparrow$} \\
    \midrule
    RXD &  & 50.0 & 55.1 & 17.6 \\
    PAD &  & 56.3 & 70.1 & 57.1 \\
    \midrule
    \multirow{5}{*}{MRAD---Grid Size} 
        & 8×8 & 75.7 & 73.6 & 60.0 \\
        & 6×6 & 76.3 & 71.8 & 61.3 \\
        & 3×3 & 72.1 & 68.1 & 61.3 \\
        & 2×2 & 65.6 & 65.1 & 59.2 \\
        & No grids & 57.9 & 58.9 & 54.7 \\
    \midrule
    \multirow{4}{*}{MRAD---Tolerance} 
        & 25\% & 74.9 & 69.4 & 60.7 \\
        & 50\% & 74.9 & 68.3 & 61.0 \\
        & 150\% & 67.5 & 76.6 & 58.9 \\
        & 200\% & 68.5 & 67.3 & 54.9 \\
    \midrule
    \multirow{4}{*}{MRAD---Kernel Size} 
        & 5\% & 74.9 & 70.8 & 61.7 \\
        & 10\% & 74.9 & 70.0 & 62.1 \\
        & 20\% & 75.0 & 71.8 & 62.9 \\
        & 30\% & 74.9 & 70.3 & 60.1 \\
    \midrule
    \multirow{3}{*}{MRAD---Noise SD} 
        & SD 2 & 73.0 & 70.7 & 62.8 \\
        & SD 5 & 75.0 & 70.6 & 62.9 \\
        & SD 10 & 67.5 & 77.3 & 59.4 \\
    \midrule
    \textbf{Optimized} & – & 75.0 & 71.8 & 62.9 \\
    \bottomrule
\end{tabular}
}
\vspace{-5mm}
\end{table}

\vspace{-2mm}
\subsection{Discussion}
\label{subsec:discussion}

\subsubsection{Limitations of Learning-Based Methods}

Analysis of our results (\cref{subsec:results}) reveals the limitations of established anomaly detection methods within the space domain. Many leverage feature extractors pre-trained on ImageNet, a dataset whose visual characteristics differ significantly from ALLO. Consequently, these networks generate non-discriminative features when applied to ALLO data, compromising their effectiveness in learning the dataset's distribution.

In addition, learning-based anomaly detection methods are often limited by their assumptions on the distribution of normal data. Approaches based on unimodality \cite{yu_fastflow_2021} fail when anomalies closely resemble normal samples, or when the normal data distribution is multi-modal. 
Many algorithms also presume image consistency---fixed lighting and viewpoints typical of industrial inspection tasks. On datasets like ALLO, where lighting and viewing conditions vary and misalignments are common, anomaly scores become unreliable. 
Feature-reconstruction models such as Dinomaly \cite{Guo_2025_CVPR}, leverage vision foundation models to learn rich features and achieve the strongest learning-based results on ALLO. 
Likewise, normalizing flow methods (FastFlow \cite{yu_fastflow_2021}, UFlow \cite{tailanianUFlowUShapedNormalizing2024a}) better model complex distributions and reduce false positives.
Nonetheless, even these advanced methods struggle to detect anomalies that closely resemble the space station, particularly under variable illumination or perspective.

Ultimately, learned methods struggle on ALLO because they attempt to model a single global distribution of non-anomalous features across all views and lighting. In contrast, MRAD analyzes each query image individually via direct reference comparison, eliminating the need for global modeling and enabling robust anomaly localization regardless of lighting or viewing variation.

\subsubsection{Statistical (Learning-Free) Anomaly Detection}

Further analysis of our results demonstrates how a non-learning algorithm can be used for anomaly detection and which components most strongly affect its performance. The value of a reference image is evident in the performance gap between RXD and PAD: computing background statistics from reference rather than query images increases the P.AP of RXD by 39.5\%. By comparing RXD scores between the query and reference images, MRAD attenuates the impact of lighting variations and relaxes the Gaussian pixel-distribution assumption of RXD. Combined with noise removal and region growing to suppress false positives and cluster anomalous pixels, MRAD yields a 5.8\% increase in P.AP and an 18.7\% improvement in I.AUROC over PAD.

The ablation studies identify grid size and region-growing tolerance as the dominant parameters. Increasing the grid size beyond the optimized 4 $\xtimes$ 4 configuration degrades both pixel-level and image-level performance, as large grids favor global context at the expense of fine-grained anomalies. In contrast, smaller grids exploit the reference image more effectively by emphasizing local deviations. Region growing tolerance governs how anomaly scores are aggregated: lower tolerances cause a modest decrease in P.AP with little effect on I.AUROC, while higher tolerances rapidly degrade performance by incorrectly including non-anomalous pixels. For noise handling, kernel size has a larger influence than the Gaussian standard deviation; larger kernels risk removing truly anomalous pixels, whereas smaller kernels fail to suppress background fluctuations. The optimized configuration (a 4 $\xtimes$ 4 grid, tolerance of 50 MRAD score, kernel size of 20\%, and standard deviation of 5) provides a practical balance between localization sensitivity and robustness.

These results illustrate how MRAD’s simple statistical formulation and few tunable parameters yield interpretable behavior and straightforward adaptation to different operational scenarios. Unlike deep learning approaches, which typically behave as black boxes and often require retraining, MRAD’s explicit use of background statistics produces predictable responses to changes in illumination and scene content. Moreover, this statistical approach avoids the data dependence of deep learning: acquiring large on-orbit datasets is prohibitively expensive, and training on synthetic imagery introduces a sim-to-real gap that can undermine on-orbit performance. MRAD sidesteps these issues by generating a reference image at inference time from a known CAD model and pose estimate.

This design also introduces limitations. Because MRAD does not learn from data, its performance is constrained by the fidelity of the CAD model, the quality of the rendered reference image, and the validity of its statistical assumptions. 

Finally, while the present analysis focuses on free-floating anomalies, since MRAD operates on a per-image basis with a rendered reference, the same framework could naturally be extended to inspection tasks such as detecting structural damage or surface degradation.

\section{Conclusion}
\label{sec:conclusion}

In this paper, we introduced ALLO, the first open-source anomaly detection dataset with synthetic lunar-orbit imagery, and established a benchmark for space-based anomaly detection by evaluating state-of-the-art methods.
We also proposed MRAD, a statistical algorithm that leverages known camera poses and a CAD model to render anomaly-free reference images and detect statistical deviations. On ALLO, MRAD achieves state-of-the-art performance, outperforming existing methods in pixel-level AP, FPR95, and image-level AUROC.

While MRAD achieves strong results, its reliance on accurate CAD models and runtime rendering introduces practical constraints. Learning-based methods, though potentially more computationally efficient, currently fail to generalize to the varied lighting and geometric conditions of the space domain. A promising direction for future work is to develop novel learning strategies, where models are trained to discriminate between normal and anomalous synthetic examples. By explicitly teaching a model what constitutes an anomaly in this specific domain, we could enhance its discriminative power and robustness to the extreme lighting and viewpoint variability, enabling reliable anomaly detection for future autonomous space operations.

\bibliographystyle{IEEEtran}
\vspace{-3mm}
\bibliography{refs_cleaned,IEEEtranBSTCTL}

\end{document}